\documentclass[sigconf]{acmart}

\usepackage{algorithm}
\usepackage{algorithmic}
\usepackage{amsfonts}
\usepackage{amsmath}
\usepackage{xcolor}
\usepackage{booktabs}
\usepackage[most]{tcolorbox}
\usepackage{calligra}

\AtBeginDocument{%
  }
\copyrightyear{2026}
\acmYear{2026}
\setcopyright{cc}
\setcctype{by}
\acmConference[KDD '26]{Proceedings of the 32nd ACM SIGKDD Conference on Knowledge Discovery and Data Mining V.2}{August 09--13, 2026}{Jeju Island, Republic of Korea}
\acmBooktitle{Proceedings of the 32nd ACM SIGKDD Conference on Knowledge Discovery and Data Mining V.2 (KDD '26), August 09--13, 2026, Jeju Island, Republic of Korea}
\acmDOI{10.1145/3770855.3817781}
\acmISBN{979-8-4007-2259-2/2026/08}

\begin{document}

\title{inversedMixup: Data Augmentation via Inverting Mixed Embeddings}

\author{Fanshuang Kong}
\email{kongfs@buaa.edu.cn}
\affiliation{
  \institution{Beihang University}
  \city{Beijing}
  \country{China}
}

\author{Richong Zhang}
\authornote{Corresponding author: zhangrichong@buaa.edu.cn.}
\email{zhangrichong@buaa.edu.cn}
\affiliation{
  \institution{Beihang University}
  \city{Beijing}
  \country{China}
}

\author{Qiyu Sun}
\email{sunqiyu@buaa.edu.cn}
\affiliation{
  \institution{Beihang University}
  \city{Beijing}
  \country{China}
}

\author{Zhijie Nie}
\email{niezhijie@buaa.edu.cn}
\affiliation{
  \institution{Beihang University}
  \city{Beijing}
  \country{China}
}

\author{Ting Deng}
\email{dengting@buaa.edu.cn}
\affiliation{
  \institution{Beihang University}
  \city{Beijing}
  \country{China}
}

\author{Chunming Hu}
\email{hucm@buaa.edu.cn}
\affiliation{
  \institution{Beihang University}
  \city{Beijing}
  \country{China}
}

\renewcommand{\shortauthors}{Fanshuang Kong et al.}

\begin{abstract}
  Mixup generates augmented samples by linearly interpolating inputs and labels with a controllable ratio. However, since it operates at the latent embedding level, the resulting samples are not human-interpretable. In contrast, LLM-based augmentation methods produce sentences via prompts at the token level, yielding readable outputs but offering limited control over the generation process. Inspired by recent advances in LLM inversion, which reconstructs natural language from embeddings and helps bridge the gap between latent embedding space and discrete token space, we propose inversedMixup, a unified framework that combines the controllability of Mixup with the interpretability of LLM-based generation. Specifically, inversedMixup aligns the output embedding space of a task-specific model with the input embedding space of an LLM, so that mixed embeddings can be reconstructed, under a controllable mixing ratio, into human-interpretable sentences. This interpretability provides the first empirical evidence of the manifold intrusion phenomenon in text Mixup. Building on this, we extend inversedMixup into a three-stage data augmentation method, and introduce a simple yet effective strategy to mitigate manifold intrusion during augmentation. Extensive experiments demonstrate the effectiveness and generalizability of our approach in both few-shot and fully supervised scenarios. Our code is available at~\url{https://github.com/pypi1412/inversedMixup}.
\end{abstract}




\maketitle

\section{Introduction}
In text understanding, low-resource scenarios are often encountered, such as comprehending data in specific domains. Data augmentation is a basic technique to address this challenge~\cite{shorten2019survey}.
Based on whether large language models (LLMs) are involved, existing augmentation approaches can be broadly categorized into traditional augmentation methods and LLM-based methods.

Among traditional augmentation methods~\cite{wei2019eda,sennrich2015improving}, Mixup~\cite{zhang2017mixup} is widely used to generate augmented samples by interpolating inputs and labels with a controllable ratio. In text augmentation, Mixup is typically applied at the embedding level. Although the interpolation results enjoy a delicate degree of controllability through the adjustment of the mixing ratio, the outcomes in the latent space lack human interpretability, resulting in synthesized sentences that are not easily understandable and may have inconsistent semantics.

LLM-based augmentation methods typically rely on prompting techniques to activate the knowledge acquired by LLMs during pretraining, thereby generating augmented samples that benefit downstream task-specific models~\citep{li-etal-2023-synthetic,gao2023selfguided,yu2023large,cui-wang-2024-ada}.
Unlike embedding-level interpolation methods, LLM-based approaches directly synthesize sentences at the token level, offering high human interpretability of the generated results. However, since LLMs are primarily driven by prompt-based instructions rather than explicit model structural constraints, the augmentation process exhibits limited controllability over the generated outputs.

\begin{figure}
    \centering
    \includegraphics[width=0.975\columnwidth]{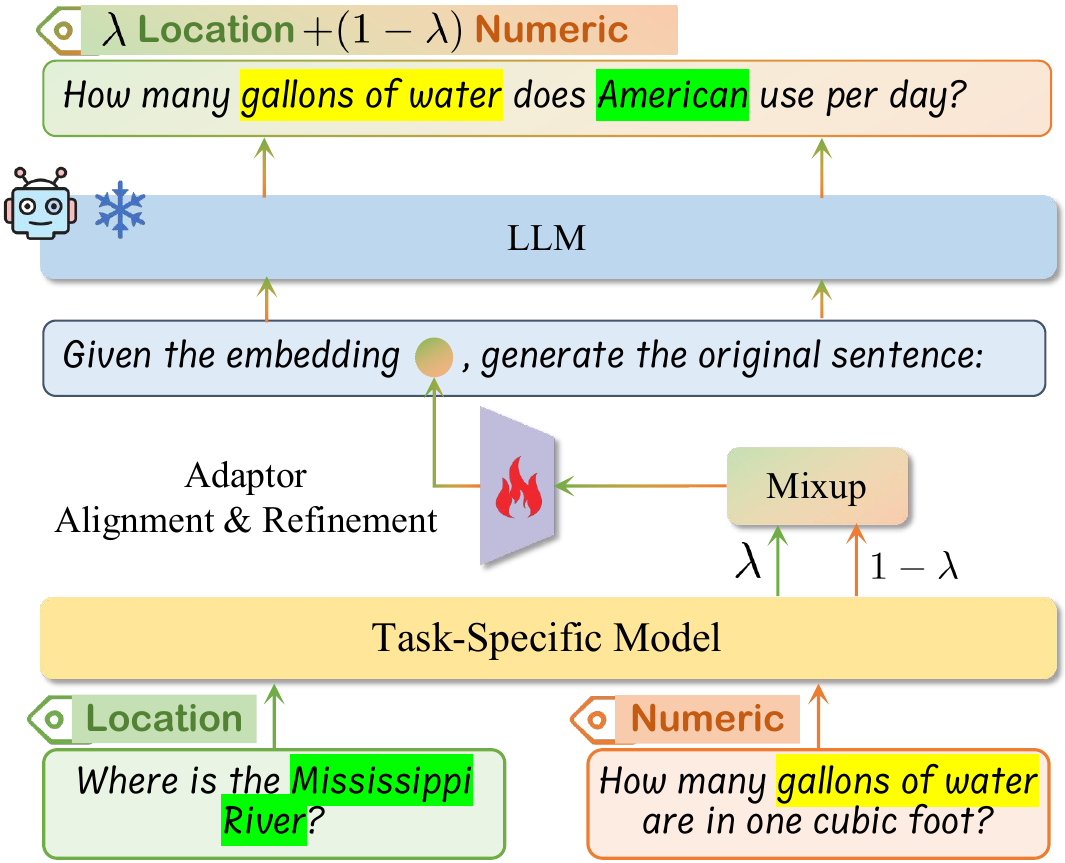} 
    \caption{Illustration of inversedMixup. Mixed embedding with a ratio $\lambda$ between a `Location'-labeled sentence (green) and a `Numeric'-labeled sentence (orange) is inverted by the LLM.
    The reconstructed sentence incorporates semantics from both source inputs.
    }
    \label{fig:motivation}
\end{figure}

Comprehensively, while Mixup enables precise controllability through explicit interpolation at the embedding level, it lacks human interpretability in the synthesized sentences. In contrast, LLM-based methods generate semantically rich and interpretable outputs, yet their token-level generation process remains difficult to control. This contrast highlights the need for a unified approach that bridges Mixup-based embedding-level interpolation and LLM-based token-level generation.

Recently, LLM inversion has emerged as a promising approach to bridge the gap between the embedding level and the token level for text. In LLM inversion, a sentence embedding is treated as a soft token input to a large language model, which leverages its generative capabilities to reconstruct sentences from the embedding~\cite{carlini2021extracting, chen2024text}. Several studies~\cite{tennenholtz2023demystifying, zhang2024map2text} have shown that this process can produce fluent and semantically consistent reconstructions. This demonstrates that LLM inversion can effectively connect continuous embedding-level latent vectors with discrete token-level sequences, establishing a direct mapping from latent semantic representations to human-interpretable text.

Building upon this insight, we propose inversedMixup, which integrates Mixup with LLM inversion to make text Mixup both controllable and human-interpretable. Specifically, Mixup operates at the embedding level to produce controllable and diverse interpolated embeddings, while LLM inversion maps these embeddings back to human-interpretable sentences.
An illustration of the inversedMixup framework is shown in Figure~\ref{fig:motivation}. A task-specific model first generates sentence embeddings, which are then interpolated using Mixup. The resulting mixed embeddings are then transformed into the LLM’s input space through an adaptor, allowing the LLM to reconstruct the corresponding sentences from these mixed embeddings.

In this way, inversedMixup provides a new avenue for observing the manifold intrusion phenomenon in text Mixup.
Manifold intrusion~\citep{verma2019manifold,guo2020nonlinear} is a common issue in Mixup, where the interpolated input does not align with the interpolated label. While this problem has been well studied in computer vision, where inputs are human-interpretable, it remains underexplored in text due to the discrete nature of language. Since interpolation must be performed in latent embedding spaces, the resulting embeddings are often difficult to interpret, making intrusion hard to detect and mitigate. By inverting each mixed embedding back into a readable sentence, our framework removes this barrier and, to our knowledge, makes manifold intrusion in text Mixup directly observable for the first time. Examining the inversion results of mixed embeddings (shown in Figure~\ref{fig:case}), we find that manifold intrusion is prevalent during the text Mixup process.

Additionally, we extend inversedMixup into a data augmentation method. Specifically, we propose a unified three-stage alignment training framework: adaptor alignment with unlabeled data, adaptor refinement via supervised warming-up, and inverting mixed embeddings.
In the adaptor alignment phase, the alignment between the task-specific model and the LLM is trained using a large open-domain corpus of widely available unlabeled data. This step ensures a general alignment between the embedding spaces of the two models.
In the adaptor refinement phase, the task-specific model and the adaptor undergo iterative fine-tuning using labeled task-specific data. This process refines the alignment and injects task information into the adaptor, analogous to the fine-tuning stage commonly used for pre-trained models.
Finally, in the inverting mixed embedding phase, latent sentence embeddings are mixed and then reconstructed into augmented samples, which are subsequently used to further train the task-specific model.
Through this three-stage training process, the LLM can accurately reconstruct mixed embeddings with a controllable ratio into human-interpretable sentences, resulting in high-quality augmented samples.
To mitigate the observed intrusion, we introduce a simple and practical solution that assigns a new label to each generated sentence using the LLM itself. Experiments (reported in Figure~\ref{fig:soft_hard}) demonstrate that this approach can effectively reduce the manifold intrusion phenomenon.

In short, our paper makes the following contributions:

\begin{itemize}
    \item We propose inversedMixup, which aligns a task-specific model with an LLM and inverts mixed embeddings to make text Mixup both controllable and human-interpretable. This interpretability lets us directly observe manifold intrusion in text Mixup, providing the first empirical evidence of this phenomenon in the text domain.
    \item Building on this observation, we extend inversedMixup into a simple yet effective three-stage text augmentation method that mitigates manifold intrusion and enhances task-specific model performance.
    \item Experiments across diverse datasets and learning setups demonstrate the effectiveness and generalizability of inversedMixup in both few-shot and fully supervised scenarios.
\end{itemize}

\section{Related Work}

\subsection{Mixup}
Mixup~\cite{zhang2017mixup} is a widely adopted data augmentation technique that generates virtual training samples by taking convex combinations of input pairs and their corresponding labels. Mixup has been shown to improve generalization~\cite{zhang2017mixup}, smooth decision boundaries~\cite{verma2019manifold}, and enhance robustness against adversarial attacks~\cite{pang2019mixup}. 

Due to the discrete nature of text, directly applying Mixup at the token level is impractical. As a result, several studies have explored Mixup in continuous latent spaces, particularly within neural language model embeddings~\citep{li2024simple,xu2024probabilistic}. For instance, \citet{guo2019augmenting} and \citet{verma2019manifold} investigate interpolation in word, sentence embeddings, or hidden state embedding, followed by classification with soft labels based on the interpolation weights. These approaches show that Mixup in latent spaces can generate semantically meaningful intermediate embeddings, improving performance in classification and domain generalization tasks~\cite{wu_inkpen_el-roby_2020}. Other variants, such as MixText~\cite{mixtext}, DM-ADA~\cite{xu2020adversarial} or contrastive Mixup~\cite{kim2020mixco}, further explore the benefits of interpolated embeddings for regularizing model behavior. However, most methods remain limited to embedding spaces and do not explicitly recover interpolated examples in natural language, hindering their interpretability and wider applicability.

\subsection{LLM Inversion}
LLM inversion investigates the capability of large language models to reconstruct input text or semantic content from intermediate embeddings. This line of research is closely related to studies on model interpretability, privacy leakage, and knowledge localization in LLMs~\cite{carlini2021extracting, morris2023language, chen2024text, zhuang2024understanding, li2023sentence}.

Recent studies delve deeper into the potential of LLMs as decoders for latent semantic embeddings. For example, \citet{tennenholtz2023demystifying} treats LLMs as interpreters of embeddings, reconstructing text from embeddings to understand their content. \citet{zhang2024map2text} further reveals that semantically similar embeddings tend to reconstruct sentences with higher coherence, highlighting the potential of LLMs to generate meaningful content from latent embeddings. This suggests that LLMs can be powerful generative tools for decoding intermediate embeddings, offering new opportunities for text augmentation.

\section{inversedMixup}
\begin{figure*}[h]
    \centering
    \includegraphics[width=0.975\linewidth]{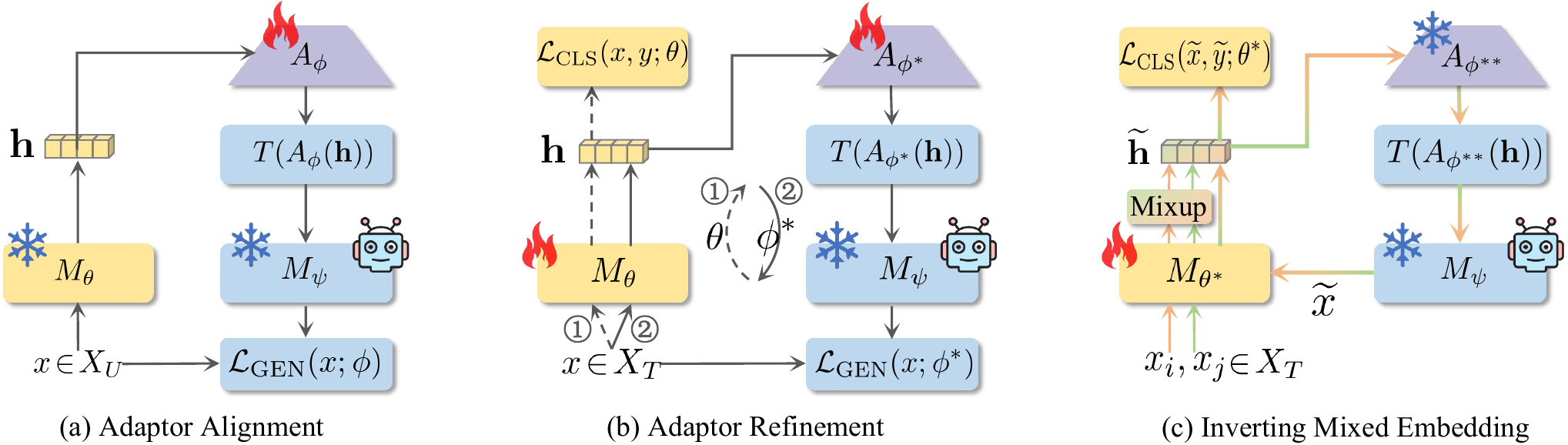}
    \caption{
    Overview of the proposed inversedMixup framework.
    (a) Adaptor alignment with unlabeled data aligns the task-specific model $M_\theta$ with the LLM $M_\psi$ by training the adaptor $A_\phi$ on large-scale unlabeled data.
    (b) Adaptor refinement further fine-tunes $M_\theta$ and $A_{\phi^*}$ using labeled target-task data to enhance alignment and task adaptability.
    (c) Inverting mixed embeddings mixes embeddings from the aligned $M_{\theta^*}$, transforms the mixed embedding into the LLM space via $A_{\phi^{**}}$, and leverages the LLM to invert the mixed embedding into augmented samples $\widetilde{x}$.
    }
    \label{fig:model}
\end{figure*}

The inversedMixup is a text augmentation framework that generates diverse samples by reconstructing mixed embeddings. It comprises three stages:
(1) Adaptor alignment with unlabeled data, which aligns the embedding spaces of a task-specific model and LLM via adaptor training on large-scale unlabeled data;
(2) Adaptor refinement via supervised warming-up, which further tunes the adaptor and the task-specific model using labeled target-task data;
(3) Inverting mixed embedding, which first mixes the embeddings generated by the task-specific model, then transforms the mixed embedding into the LLM embedding space through the aligned adaptor, and finally utilizes the LLM to invert the mixed embedding to produce augmented samples.
The augmented samples, along with the original task data, are jointly fine-tuned to enhance the task-specific model, leading to improved performance and better generalization on the target task. Figure~\ref{fig:model} illustrates the three stages of inversedMixup.

\subsection{Notations}
Given a target task, let $(X_T, Y_T)$ denote all labeled data, where each sample consists of an input $x$ and its label $y$, with corresponding one-hot label vector denoted as $\mathbf{y}$.
Typical data augmentation methods synthesize valuable additional samples based on the distribution of $(X_T, Y_T)$ to train a more effective task-specific model $M_\theta$ (with parameters $\theta$).
In inversedMixup, we further utilize widely available open-domain unlabeled data $X_U$ to train a learnable adaptor $A_\phi$, which aligns the output space of the task-specific model $M_\theta$ with the input space of a generative LLM $M_\psi$.
By leveraging this strong alignment and performing inversion on mixed embeddings, we generate augmented samples that substantially enhance the performance of $M_\theta$ on the target task.

In our experiments, we evaluate the effectiveness of alignment and augmentation under both few-shot and fully-supervised settings, based on the number of labeled examples. Additionally, we consider a special case where $X_T$ is sampled directly from $X_U$, in which the task-specific information is expected to be best preserved due to domain consistency.

\subsection{Adaptor Alignment with Unlabeled Data}
As demonstrated in~\citep{chen2024text, li2023sentence}, with sufficiently extensive training, the LLM $M_\psi$ is capable of effectively reconstructing the original sentence $x$ from its embedding, a task named the LLM inversion.
In this paper, we aim to leverage the powerful generative capability of LLM $M_\psi$ to reconstruct the embeddings produced by a task-specific model $M_\theta$. Since the output space of the task-specific model typically does not align with the input space of the LLM, we introduce a learnable adaptor $A_\phi$ to effectively bridge the two spaces.
Ideally, this alignment can be achieved by training the adaptor on large-scale open-domain unlabeled data using the self-supervised loss of the language model.

Formally, given an input sentence $x$ drawn from the unlabeled dataset $X_U$, the task-specific model $M_\theta$ generates its corresponding sentence embedding $\mathbf{h}$, formally denoted as:
\begin{equation}
    \mathbf{h} = M_\theta(x).
\end{equation} 

\newtcolorbox{promptbox}{
  colback=gray!10!white, colframe=black, sharp corners,
  boxrule=0.3mm, top=5pt, bottom=5pt, left=5pt, right=5pt, breakable
}

Subsequently, the embedding $\mathbf{h}$ is mapped into the input embedding space of $M_\psi$ through the adaptor $A_\phi$. Similar to popular multimodal or graph alignment approaches, $A_\phi(\mathbf{h})$ serves as a soft token that, when combined with an appropriate prompt, guides the LLM to reconstruct the original sentence $x$ corresponding to the embedding.
Specifically, we define the input prompt as a function $T$ based on $A_{\phi}(\mathbf{h})$, i.e., $T(A_{\phi}(\mathbf{h}))$, and the prompt is defined as:

\begin{promptbox}
    Given the embedding [$A_{\phi}(\mathbf{h})$], generate the original sentence: [$x$] .
\end{promptbox}

The optimization objective of the LLM is to minimize the negative log-likelihood of the sequence \( x \) conditioned on the adapted embedding \( A_{\phi}(\mathbf{h}) \). This can be concisely and unambiguously expressed as:
\begin{equation}
    \mathcal{L}_{\text{GEN}}(x; \phi) = - \log P_{\psi}\big(x \mid A_{\phi}(\mathbf{h})\big).
\end{equation}
Noted, during this training phase, only the adaptor parameters $\phi$ are updated, while the parameters of both the task-specific model $M_\theta$ and the LLM $M_\psi$ remain fixed.
That is, the adaptor is optimized over the entire unlabeled dataset $X_U$ to obtain the optimal parameters $\phi^*$ by minimizing the generation loss:
\begin{equation}
    \phi^* = \mathop{\arg \min}\limits_{\phi} \sum_{x \in X_U} \mathcal{L}_{\text{GEN}}(x;\phi).
\end{equation}

\subsection{Adaptor Refinement via Supervised Warming-up}
While the initial alignment enables the large model $M_\psi$ to roughly interpret the embedding space of $M_\theta$, this mapping is often insufficient when applied to specific tasks. As the task-specific model is usually further fine-tuned on task data, its embedding distribution inevitably shifts, resulting in a mismatch with the previously trained adaptor $A_{\phi^*}$. Moreover, since $A_{\phi^*}$ is trained solely on open-domain unlabeled data, it lacks the task-specific inductive bias necessary for effective adaptation. To bridge this gap, we introduce a supervised warming-up phase, during which the task-specific model $M_\theta$ and the adaptor $A_{\phi^*}$ are iteratively fine-tuned on labeled data from the target task. This iterative update strategy allows $M_\psi$ to continuously align well with the task-specific model while simultaneously incorporating task information into the alignment process.

Formally, given a labeled training sample $(x, y)$ from the task-specific dataset $(X_T, Y_T)$, the task-specific model $M_\theta$ is first applied to obtain the sentence embedding $\mathbf{h}$, based on which the predicted label distribution is computed as (take the classification task as an illustrative example):
\begin{equation}
    \hat{\mathbf{y}} = f_{\mathbf{h} \mapsto \mathbf{y}}(M_\theta(x)),
\end{equation}
where \(\hat{\mathbf{y}} \in \mathbb{R}^K\) denotes the predicted logits over \(K\) predefined classes. The classification head \(f_{\mathbf{h} \mapsto \mathbf{y}}\) can be instantiated either as a lightweight multi-layer perceptron (MLP) or as a prompt-based decoding module for the target task.

To train the task-specific model $M_\theta$ under supervision, we minimize the standard cross-entropy loss between the predicted logits $\hat{\mathbf{y}}$ and the ground-truth label $y$ (corresponding one-hot label vector $\mathbf{y}$), formulated as:
\begin{equation}
    \mathcal{L}_{\text{CLS}}(x, y; \theta) = -\sum_{k=1}^{K}\mathbf{y}_k\log \hat{\mathbf{y}}_k.
\end{equation}
And the optimal task-specific model parameters $\theta^*$ could be obtained by minimizing the classification loss over the entire labeled dataset $(X_T, Y_T)$:
\begin{equation}
    \theta^* = \mathop{\arg\min}_{\theta} \sum_{(x,y) \in (X_T, Y_T)} \mathcal{L}_{\text{CLS}}(x, y; \theta).
\end{equation}

After obtaining the optimal parameters $\theta^*$, the embedding distribution of the task-specific model inevitably shifts, rendering the adaptor $A_{\phi^*}$ suboptimal. To restore alignment with the large model $M_\psi$, the adaptor $A_{\phi^*}$ must be refined to accommodate the updated embeddings and adapt to task data $X_T$. Generally, the updated adaptor parameters $\phi^{**}$ are obtained by:
\begin{equation}
    \begin{aligned}
        \phi^{**} & = \mathop{\arg \min}\limits_{\phi^*} \sum_{x \in X_T} \mathcal{L}_{\text{GEN}}(x; \phi^*) \\
        & = \mathop{\arg \min}\limits_{\phi^*} - \sum_{x \in X_T} \log P_\psi(x \mid A_{\phi^*}(M_{\theta^*}(x))).
    \end{aligned}
\end{equation}
To ensure robust alignment, $\theta^*$ and $\phi^{**}$ are optimized iteratively during the refinement phase. In general, through iterative adaptor refinement, $M_{\theta^*}$ and $M_\psi$ can be better aligned, conditioned on the task-specific data $X_T$. Moreover, the more samples available in $X_T$, the better the alignment performance.

\subsection{Inverting Mixed Embedding}
Following the above procedure, the LLM $M_\psi$ should be able to reconstruct the original sentence given an embedding produced by the task-specific model $M_{\theta^*}$. Prior works~\citep{tennenholtz2023demystifying,zhang2024map2text} have shown that a well-trained adaptor, when paired with a powerful large model, can reliably recover the original sentence from its corresponding embedding while preserving the underlying semantics.
In the context of text augmentation, such a property opens up an intriguing possibility: if two embeddings are interpolated (e.g., via Mixup) in the embedding space, the large model can potentially decode the interpolated embedding into a synthetic sentence whose semantics lie between those of the two original inputs. Ideally, if the large model possesses sufficient generative capacity, this decoding process can yield high-quality and semantically meaningful augmented samples for the specific task.

Building on this intuition, we aim to generate synthetic samples with diverse semantic labels by mixing embeddings at varying ratios. These interpolated embeddings are then decoded by the large model $M_\psi$ into synthetic sentences that enrich the training data, thereby enhancing the robustness and generalization of the task-specific model $M_{\theta^*}$.
Formally, given two labeled examples $(x_i, y_i)$ and $(x_j, y_j)$ from the task dataset $(X_T, Y_T)$, we first compute their sentence embeddings using the refined aligned model $M_{\theta^*}$:
\begin{equation}
    \mathbf{h}_i = M_{\theta^*}(x_i), \quad 
    \mathbf{h}_j = M_{\theta^*}(x_j).
\end{equation}
These embeddings are then interpolated using the vanilla Mixup strategy:
\begin{equation}
    \widetilde{\mathbf{h}} = \lambda \mathbf{h}_i + (1 - \lambda) \mathbf{h}_j, \quad 
    \lambda \sim \text{Beta}(\alpha, \alpha),
\end{equation}
where $\alpha$ is a hyperparameter that controls the ratio of interpolation.
The resulting mixed embedding $\widetilde{\mathbf{h}}$ is then passed through the adaptor $A_{\phi^{**}}$, and decoded by the LLM $M_\psi$ to generate a synthetic sentence $\widetilde{x}$:
\begin{equation}
    \widetilde{x} = M_\psi(T(A_{\phi^{**}}(\widetilde{\mathbf{h}}))).
\end{equation}
Following the vanilla Mixup principle, where linear interpolation is applied to both input $x$ and its label $y$ to maintain semantic consistency, we also apply label mixup in a similar fashion:
\begin{equation}
    \widetilde{\mathbf{y}} = \lambda \mathbf{y}_i + (1 - \lambda) \mathbf{y}_j.
\end{equation}
The synthetic pair $(\widetilde{x}, \widetilde{\mathbf{y}})$, or equivalently $(\widetilde{x}, \widetilde{y})$ when there is no ambiguity, can be incorporated into the training set to enhance the task-specific model's performance.
Generally, the Mixup ratio $\lambda$ can be flexibly adjusted, enabling the generation of a potentially unlimited number of high-quality synthetic samples. These samples can be used either to further refine the alignment adaptor or to enhance the training of the task-specific model.

Let the set of generated samples be denoted as $(\widetilde{X}, \widetilde{Y})$. By combining them with the original labeled dataset $(X_T, Y_T)$, we construct an augmented training set $(X_{\text{aug}}, Y_{\text{aug}}) = (X_T, Y_T) \cup (\widetilde{X}, \widetilde{Y})$.
The final objective for the task-specific model $M_{\theta^*}$ is then updated as:
\begin{equation}
    \min_{\theta^*} \sum_{(x, y) \in (X_{\text{aug}}, Y_{\text{aug}})} \mathcal{L}_{\text{CLS}}(x, y; \theta^*).
\end{equation}
With this, the inversedMixup framework completes its augmentation process by generating synthetic samples and leveraging them to train the task-specific model.

\begin{figure}
    \centering
    \begin{tabular}{@{}c@{\hspace{1mm}}c@{}}
        \includegraphics[width=0.47\linewidth]{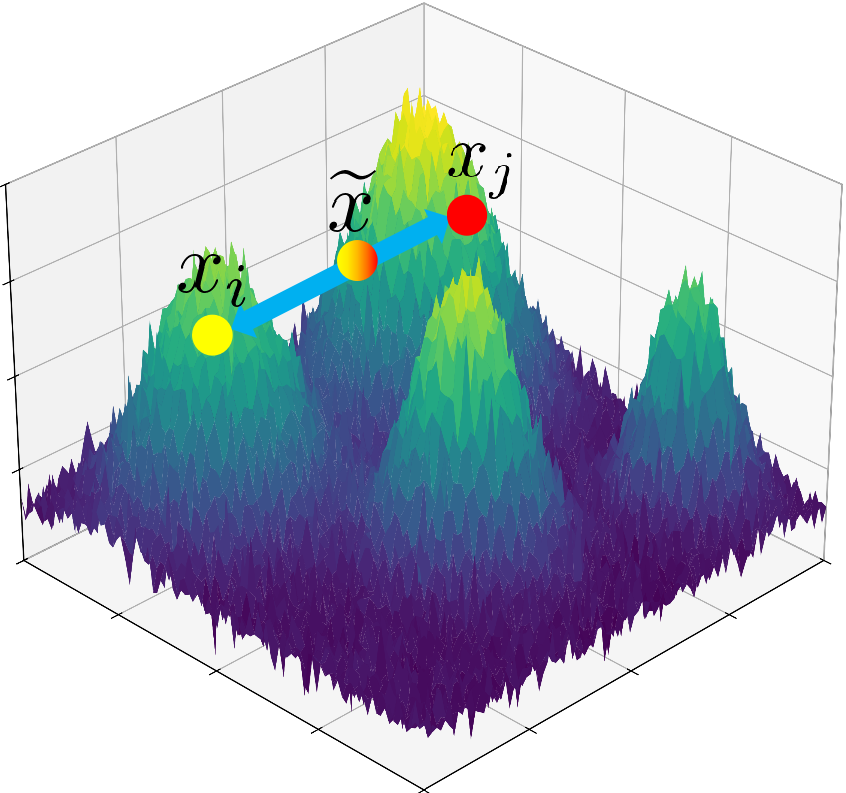} & \includegraphics[width=0.47\linewidth]{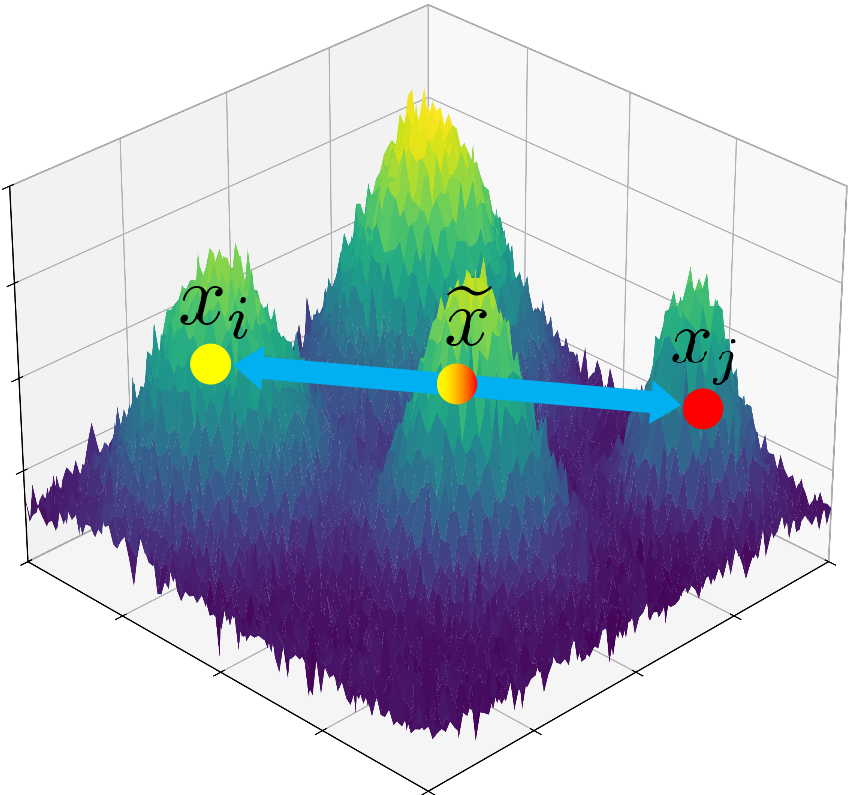} \\
        (a) Inside Intrusion & (b) Outside Intrusion
    \end{tabular}
    \caption{`Inside' refers to cases where the interpolated sample $\widetilde{x}$ is assigned a label that matches either $x_i$ or $x_j$, indicating it still lies within the original class manifold. `Outside' refers to cases where $\widetilde{x}$ receives a label unrelated to either original class, reflecting a deviation from the class manifold.}
    \label{fig:intrusion}
\end{figure}

\subsection{Manifold Intrusion: Existence and Alleviation}
\label{sec:intrusion}
Despite the effectiveness of Mixup in generating diverse training samples, it often suffers from manifold intrusion—a phenomenon where the interpolated embedding fails to semantically correspond to the convex combination of the original labels.
In this paper, we categorize manifold intrusion based on whether the actual label of the interpolated sample belongs to the original input labels, dividing it into `Inside' and `Outside' intrusion. Specifically, `Inside' denotes cases where the mixed sample’s label comes from one of the original labels, while `Outside' indicates that the mixed sample’s label does not belong to either of the original labels.
Figure~\ref{fig:intrusion} illustrates the `Inside' and `Outside' intrusion issue.
Due to the human-readability and strong interpretability of input images and their interpolations, such cases are easily observable in vision tasks.
However, they are notoriously difficult to detect in text, primarily because of the lack of interpretability in latent embeddings.

Under our inversedMixup framework, a key advantage lies in its ability to reconstruct human-readable sentences from mixed embeddings, providing a level of interpretability previously unattainable.
Leveraging this interpretability, we manually inspect the labels of the inverted samples to evaluate whether manifold intrusion occurs in NLP tasks.
Specifically, we perform interpolation under various mixing ratios, with representative cases shown in Figure~\ref{fig:case}. Detailed description of the Figure refers to Section~\ref{sec:existence}. 
The results reveal that manifold intrusion occurs widely across text augmentation, regardless of the interpolation ratio.

After verifying the existence of the manifold intrusion phenomenon, we propose a simple yet effective strategy to alleviate this issue: leveraging the LLM to assign a hard one-hot label to each reconstructed sentence. A comparison between the interpolated soft labels and the LLM-predicted hard labels is presented in Figure~\ref{fig:soft_hard} (detailed in Section~\ref{sec:alleviate}). The results demonstrate that this strategy mitigates the effects of manifold intrusion and further improves performance on downstream tasks.

\section{Experiments}
\begin{table*}[htb]
\centering
\caption{Main results of inversedMixup and comparable baselines. `K' denotes the number of few-shot training samples per class, while `All' indicates the fully supervised setting where the entire training set is used.
}
\setlength{\tabcolsep}{4.5pt}
\begin{tabular}{c|cccc|cccc|cccc}
\toprule
 & \multicolumn{4}{c|}{Yahoo} & \multicolumn{4}{c|}{TREC} & \multicolumn{4}{c}{AG News} \\
\midrule
 & K=1 & K=5 & K=10 & All & K=1 & K=5 & K=10 & All & K=1 & K=5 & K=10 & All \\
\midrule
Base
 & 47.7{\scriptsize $\pm$3.2} & 58.9{\scriptsize $\pm$1.6}
 & 62.1{\scriptsize $\pm$1.3} & 69.1{\scriptsize $\pm$0.7} 
 & 36.8{\scriptsize $\pm$6.6} & 59.2{\scriptsize $\pm$6.1}
 & 72.2{\scriptsize $\pm$5.3} & 96.0{\scriptsize $\pm$0.4} 
 & 70.1{\scriptsize $\pm$4.7} & 80.7{\scriptsize $\pm$2.9}
 & 83.6{\scriptsize $\pm$1.1} & 90.0{\scriptsize $\pm$0.6} \\
\midrule
\multicolumn{13}{c}{Traditional Text Augmentation Method} \\
\midrule
 EDA 
 & 48.4{\scriptsize $\pm$4.7} & 60.8{\scriptsize $\pm$1.7}
 & 62.6{\scriptsize $\pm$1.1} & 68.0{\scriptsize $\pm$0.7} 
 & 36.7{\scriptsize $\pm$8.8} & 51.8{\scriptsize $\pm$7.6}
 & 66.6{\scriptsize $\pm$4.2} & 94.7{\scriptsize $\pm$0.4}
 & 68.1{\scriptsize $\pm$4.2} & 80.5{\scriptsize $\pm$2.9}
 & 83.8{\scriptsize $\pm$2.0} & 89.6{\scriptsize $\pm$0.2} \\
 BT
 & 48.2{\scriptsize $\pm$2.2} & 58.8{\scriptsize $\pm$2.5}
 & 62.1{\scriptsize $\pm$1.9} & 68.7{\scriptsize $\pm$1.1} 
 & 38.7{\scriptsize $\pm$8.4} & 58.2{\scriptsize $\pm$8.0}
 & 69.4{\scriptsize $\pm$5.2} & 96.0{\scriptsize $\pm$0.6}
 & 70.5{\scriptsize $\pm$7.3} & 81.0{\scriptsize $\pm$1.7}
 & 83.7{\scriptsize $\pm$1.2} & 89.6{\scriptsize $\pm$0.6} \\
 Textsmooth
 & 46.1{\scriptsize $\pm$3.0} & 56.2{\scriptsize $\pm$3.8}
 & 61.0{\scriptsize $\pm$1.8} & 68.9{\scriptsize $\pm$0.9}
 & 37.7{\scriptsize $\pm$7.0} & 57.9{\scriptsize $\pm$8.2}
 & 71.1{\scriptsize $\pm$6.9} & 95.8{\scriptsize $\pm$0.9}
 & 65.1{\scriptsize $\pm$6.4} & 74.5{\scriptsize $\pm$4.5}
 & 80.3{\scriptsize $\pm$6.0} & 89.9{\scriptsize $\pm$0.6} \\
 AWD
 & 43.1{\scriptsize $\pm$3.1} & 59.0{\scriptsize $\pm$1.2}
 & 61.4{\scriptsize $\pm$1.6} & 66.9{\scriptsize $\pm$0.8}
 & 36.2{\scriptsize $\pm$8.0} & 59.2{\scriptsize $\pm$6.6}
 & 70.8{\scriptsize $\pm$5.2} & 95.8{\scriptsize $\pm$1.0}
 & 60.0{\scriptsize $\pm$8.0} & 75.3{\scriptsize $\pm$5.5}
 & 81.6{\scriptsize $\pm$2.0} & 89.0{\scriptsize $\pm$0.8} \\
 Mixup 
 & 51.6{\scriptsize $\pm$3.5} & 61.1{\scriptsize $\pm$2.0}
 & 64.0{\scriptsize $\pm$0.9} & 67.9{\scriptsize $\pm$1.2} 
 & 37.7{\scriptsize $\pm$5.4} & 60.4{\scriptsize $\pm$7.8}
 & 73.1{\scriptsize $\pm$6.0} & 95.7{\scriptsize $\pm$1.0} 
 & 76.4{\scriptsize $\pm$2.4} & 82.5{\scriptsize $\pm$1.8}
 & 83.9{\scriptsize $\pm$1.4} & 89.0{\scriptsize $\pm$0.4} \\
\midrule
\multicolumn{13}{c}{LLM-based Text Augmentation Method }\\
\midrule
LLM-Rew 
 & 50.1{\scriptsize $\pm$2.5} & 60.6{\scriptsize $\pm$1.7}
 & 62.9{\scriptsize $\pm$1.4} & 68.7{\scriptsize $\pm$0.9} 
 & 37.3{\scriptsize $\pm$5.1} & 57.7{\scriptsize $\pm$7.0}
 & 68.9{\scriptsize $\pm$5.0} & 96.2{\scriptsize $\pm$0.5} 
 & 69.1{\scriptsize $\pm$5.8} & 80.1{\scriptsize $\pm$3.7}
 & 82.7{\scriptsize $\pm$2.4} & 89.7{\scriptsize $\pm$0.1} \\
LLM-Gen 
 & 52.9{\scriptsize $\pm$2.1} & 60.4{\scriptsize $\pm$0.9}
 & 62.3{\scriptsize $\pm$1.3} & 69.1{\scriptsize $\pm$0.8} 
 & 33.5{\scriptsize $\pm$7.1} & 56.3{\scriptsize $\pm$5.7}
 & 67.5{\scriptsize $\pm$4.2} & 95.3{\scriptsize $\pm$0.6} 
 & 72.2{\scriptsize $\pm$5.1} & 80.9{\scriptsize $\pm$3.0}
 & 82.2{\scriptsize $\pm$2.9} & 89.8{\scriptsize $\pm$0.1} \\
LLM-Mix 
 & 44.2{\scriptsize $\pm$3.6} & 57.5{\scriptsize $\pm$1.4}
 & 61.8{\scriptsize $\pm$1.3} & 69.4{\scriptsize $\pm$0.8} 
 & 35.4{\scriptsize $\pm$4.7} & 59.0{\scriptsize $\pm$6.4}
 & 70.5{\scriptsize $\pm$5.9} & 96.4{\scriptsize $\pm$1.0} 
 & 57.5{\scriptsize $\pm$4.0} & 73.9{\scriptsize $\pm$3.1}
 & 76.6{\scriptsize $\pm$3.8} & \textbf{90.3}{\scriptsize $\pm$0.1} \\
\midrule
inversedMixup 
 & \textbf{54.9}{\scriptsize $\pm$2.9} & \textbf{62.0}{\scriptsize $\pm$1.3}
 & \textbf{65.6}{\scriptsize $\pm$0.8} & \textbf{70.3}{\scriptsize $\pm$1.0} 
 & \textbf{41.6}{\scriptsize $\pm$4.6} & \textbf{61.4}{\scriptsize $\pm$6.3}
 & \textbf{74.1}{\scriptsize $\pm$4.9} & \textbf{96.5}{\scriptsize $\pm$0.7} 
 & \textbf{77.7}{\scriptsize $\pm$7.8} & \textbf{83.4}{\scriptsize $\pm$1.5}
 & \textbf{84.9}{\scriptsize $\pm$1.7} & 89.7{\scriptsize $\pm$0.4} \\
\bottomrule
\end{tabular}
\label{tab:main_results}
\end{table*}

\subsection{Experimental Setting}
\subsubsection{Datasets and Evaluation Metric}
In the adaptor alignment phase, since the training process does not require labeled data, any large-scale open-domain text corpus can be used. For our experiments, we use four widely used public datasets, Yahoo~\cite{zhang2015character}, Amazon~\cite{mcauley2013hidden}, DBpedia~\cite{dbpedia}, and Yelp~\cite{zhang2015character}, to perform the initial embedding alignment.
In the adaptor refinement and inverting mixed embedding stages, to comprehensively evaluate the applicability of our framework, we test it on two types of tasks: (1) tasks whose data were used during the alignment phase, and (2) tasks whose data were entirely unseen during alignment. Specifically, we select Yahoo as the seen dataset, and TREC~\cite{li2002learning} and AG News~\cite{zhang2015character} as unseen datasets. Detailed data statistics are shown in Appendix~\ref{app:data}. To assess the generality of our approach, we conduct experiments under both few-shot and fully supervised settings for each dataset.
For evaluation, we use accuracy as the metric to measure model performance.

\subsubsection{Implementation Details}
For the task-specific model, we adopt the widely used BERT-base-uncased~\cite{devlin2019bert}, while for the LLM, we employ the open-source LLaMA-3-8B-Instruct~\cite{dubey2024llama}. The adaptor is implemented as a lightweight single-layer MLP. The parameters are optimized using the Adam~\cite{kingma2014adam} optimizer, with a typical learning rate of 2e-5.
All experiments are conducted using 10 random seeds, and we report both the average performance and the variance across runs. Experiments are performed on NVIDIA A100 GPU. Under a moderate few-shot setting (5-shot), each epoch typically requires approximately 100 seconds for the refinement stage and 40 seconds for the embedding mixup and inversion stage.

\subsubsection{Baselines}
Data augmentation is a fundamental technique for improving model performance. 
We categorize existing methods into two main groups: traditional text augmentation methods and LLM-based text augmentation methods.
For traditional methods, we compare several widely used and effective baselines, including the classic \textbf{EDA}~\cite{wei2019eda} and \textbf{BT} (Back-Translation)~\cite{sennrich2015improving}, as well as three more robust and consistently strong methods: \textbf{Textsmooth}~\cite{wu2022text}, \textbf{AWD}~\cite{chen2023adversarial}, and \textbf{Mixup}~\cite{zhang2017mixup}.
For LLM-based methods, we consider three strategies:
\textbf{LLM-Rew}, \textbf{LLM-Gen}, \textbf{LLM-Mix}, which leverage prompt learning to respectively let the LLM rewrite existing examples, generate new ones from scratch, or directly produce mixed examples.
Details of these comparable baselines are provided in the Appendix~\ref{app:baseline}.

\begin{table}
\centering
\caption{
Wilcoxon signed-rank $p$-values comparing inversedMixup with Mixup and LLM-Mix across different numbers of few-shot samples. 
Bold values indicate statistically significant differences ($p < 0.05$).
}
\label{tab:ttest_results}
\setlength{\tabcolsep}{2pt}
\begin{tabular}{c|ccc|ccc|ccc}
\toprule
& \multicolumn{3}{c|}{Yahoo} & \multicolumn{3}{c|}{TREC} & \multicolumn{3}{c}{AG News} \\
\midrule
$10^{-2}$ & K=1 & K=5 & K=10 & K=1 & K=5 & K=10 & K=1 & K=5 & K=10 \\
\midrule
vs. Mixup
 & \textbf{0.98} & \textbf{4.20} & \textbf{0.68} &  \textbf{0.54} & 6.25 & 24.61 & 16.11 & \textbf{1.56} & 27.83 \\
vs. LLM-Mix
 & \textbf{0.20} & \textbf{0.10} & \textbf{0.20} &\textbf{0.29}  & \textbf{1.17} & \textbf{0.29} & \textbf{0.78} & \textbf{0.10} & \textbf{0.20} \\
\bottomrule
\end{tabular}
\end{table}

\begin{table*}[htb]
\centering
\caption{Effect of the adaptor refinement stage. `w/o Refinement' removes the adaptor refinement stage.
}
\setlength{\tabcolsep}{4.5pt}
\begin{tabular}{l|cccc|cccc|cccc}
\toprule
 & \multicolumn{4}{c|}{Yahoo} & \multicolumn{4}{c|}{TREC} & \multicolumn{4}{c}{AG News} \\
\midrule
 & K=1 & K=5 & K=10 & All & K=1 & K=5 & K=10 & All & K=1 & K=5 & K=10 & All \\
\midrule
inversedMixup 
 & \textbf{54.9}{\scriptsize $\pm$2.9} & \textbf{62.0}{\scriptsize $\pm$1.3}
 & \textbf{65.6}{\scriptsize $\pm$0.8} & \textbf{70.3}{\scriptsize $\pm$1.0} 
 & \textbf{41.6}{\scriptsize $\pm$4.6} & \textbf{61.4}{\scriptsize $\pm$6.3}
 & \textbf{74.1}{\scriptsize $\pm$4.9} & \textbf{96.5}{\scriptsize $\pm$0.7} 
 & \textbf{77.7}{\scriptsize $\pm$7.8} & \textbf{83.4}{\scriptsize $\pm$1.5}
 & \textbf{84.9}{\scriptsize $\pm$1.7} & \textbf{89.7}{\scriptsize $\pm$0.4} \\
w/o Refinement
 & 47.2{\scriptsize $\pm$3.1} & 57.3{\scriptsize $\pm$2.0}
 & 60.4{\scriptsize $\pm$0.9} & 68.4{\scriptsize $\pm$0.6} 
 & 30.8{\scriptsize $\pm$5.5} & 54.7{\scriptsize $\pm$2.9}
 & 65.3{\scriptsize $\pm$5.4} & 95.0{\scriptsize $\pm$0.9} 
 & 56.4{\scriptsize $\pm$6.7} & 73.9{\scriptsize $\pm$1.6}
 & 78.2{\scriptsize $\pm$1.3} & 89.6{\scriptsize $\pm$0.2} \\

\bottomrule
\end{tabular}
\label{tab:ablation}
\end{table*}

\subsection{Main Results}
Table~\ref{tab:main_results} reports the performance of inversedMixup and baseline augmentation methods on Yahoo, TREC, and AG News under few-shot (K=1, 5, 10) and fully supervised (All) settings. Overall, inversedMixup consistently achieves the best results across most settings, outperforming both traditional methods and LLM-based approaches.

From the results, we derive two key observations. First, the improvement brought by inversedMixup is especially pronounced in low-resource settings, underscoring its capability to enhance model performance when training data is limited. Second, Yahoo, which participates in both the adaptor alignment and refinement stages, exhibits the most significant performance gain. This indicates that when the embedding spaces are well aligned, inversedMixup can effectively invert the mixed embeddings and thereby serve as a powerful augmentation strategy.

\subsection{Statistical Comparison of Representative Methods}
As analyzed above, various data augmentation methods tend to yield larger improvements in few-shot settings, albeit with higher variance. To rigorously demonstrate the superiority of inversedMixup over other augmentation techniques, we employ the Wilcoxon signed-rank test~\cite{woolson2007wilcoxon} to assess the statistical significance of its results. Our comparisons focus primarily on inversedMixup versus two types of Mixup-based methods (Mixup and LLM-Mix). Following standard practice, a p-value below 0.05 indicates statistical significance. The comparison results are presented in Table~\ref{tab:ttest_results}.

From the Table, we observe that out of 18 evaluation metrics, inversedMixup achieves statistical significance on 14, providing strong evidence of its effectiveness.

\begin{figure}
    \centering
    \begin{tabular}{@{}c@{\hspace{3mm}}c@{}}
        \includegraphics[width=0.45\columnwidth]{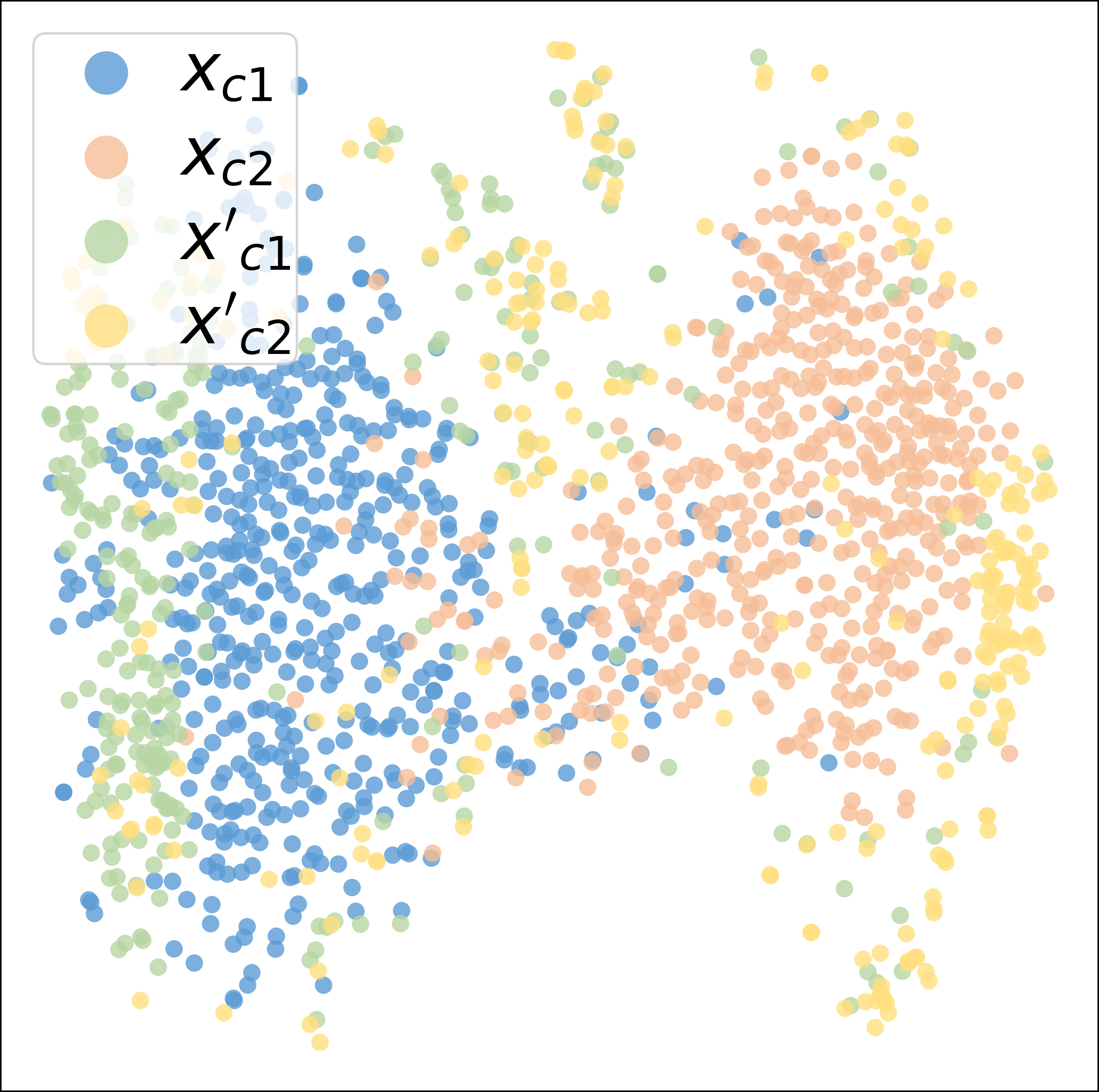} & \includegraphics[width=0.45\columnwidth]{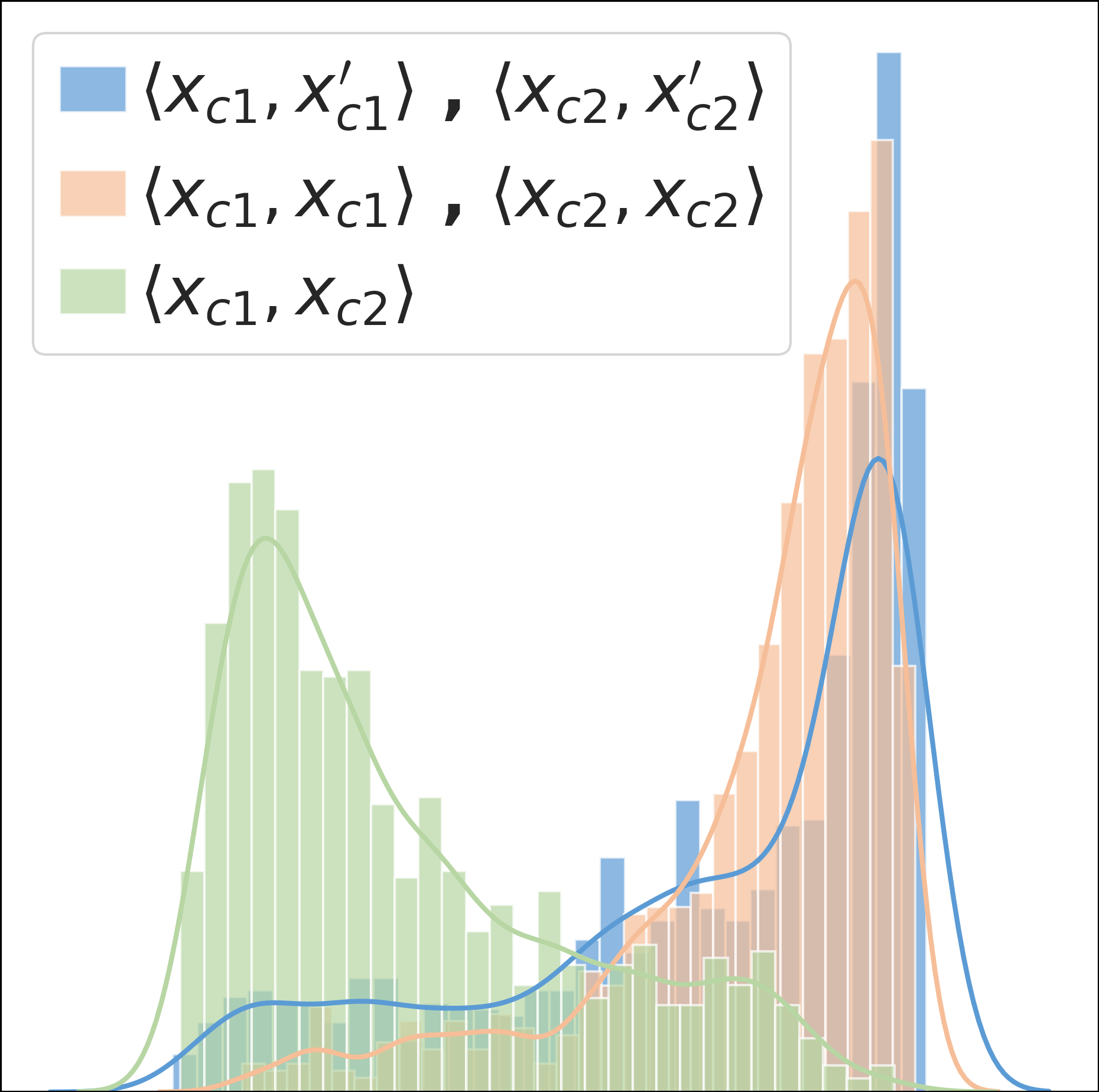} \\
        (a) T-SNE Visualization & (b) Similarity Distribution
    \end{tabular}
        \caption{(a) T-SNE visualization of the embedding of the original and reconstructed sentence. (b) The similarity distribution of different combinations of sentence pairs.}
    \label{fig:vis}
\end{figure}

\subsection{Visualization of the Inverted Embedding}

To evaluate whether the adaptor is capable of aligning embeddings, we begin by visualizing the embeddings of original and reconstructed sentences. Given a sentence $x$, we generate its reconstructed counterpart $x^{\prime}$, and obtain embeddings for both. For clarity, we select sentences from two distinct classes and apply T-SNE to visualize the embeddings of both $x$ and $x^{\prime}$. Specially, we denote the embeddings from class 1 as $x_{c1}$ and $x^{\prime}_{c1}$, and those from class 2 as $x_{c2}$ and $x^{\prime}_{c2}$. The visualization results are shown in Figure~\ref{fig:vis} (a). From the Figure, we observe that the reconstructed sentences largely preserve the characteristics of their original classes.

To further quantify the semantic consistency between original and reconstructed sentences, we compute the embedding similarity across several groups: (1) between original sentences from different classes (e.g., $\langle x_{c1},x_{c2} \rangle$), (2) between original sentences from the same class (e.g., $\langle x_{c1}, x_{c1} \rangle$, or $\langle x_{c2},x_{c2} \rangle$), and (3) between original and reconstructed sentences from the same class (e.g., $\langle x_{c1},x^{\prime}_{c1} \rangle$, or $\langle x_{c2},x^{\prime}_{c2} \rangle$). The results are shown in Figure~\ref{fig:vis} (b). From the Figure, we find that original sentence pairs from the same class have significantly higher similarity than those from different classes. Moreover, the similarity between the original and reconstructed sentences is comparable to that of same-class pairs. This shows that the reconstruction process effectively preserves semantic information and provides a solid foundation for inverting mixed embeddings.

\begin{figure}
    \centering
    \includegraphics[width=0.47\textwidth]{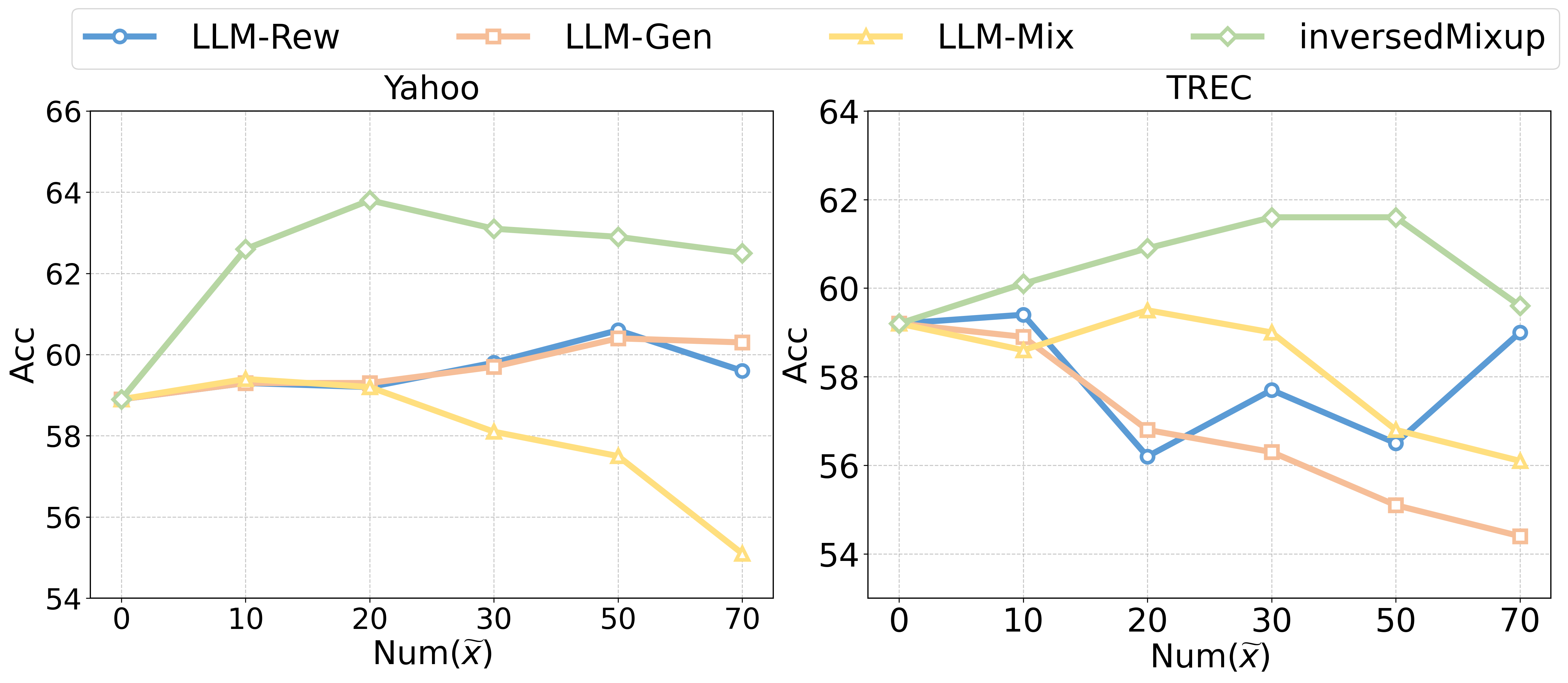}
    \caption{Comparison of different LLM-based augmentation methods with inversedMixup. $\text{Num}(\widetilde{x})$ represents the number of augmented sample.}
    \label{fig:accurancy_count}
\end{figure}

\begin{table*}[htb]
\centering
\caption{Results of inversedMixup and comparable baselines on Qwen.}
\setlength{\tabcolsep}{4.5pt}
\begin{tabular}{c|cccc|cccc|cccc}
\toprule
 & \multicolumn{4}{c|}{Yahoo} & \multicolumn{4}{c|}{TREC} & \multicolumn{4}{c}{AG News} \\
\midrule
 & K=1 & K=5 & K=10 & All & K=1 & K=5 & K=10 & All & K=1 & K=5 & K=10 & All \\
\midrule
Mixup 
 & 51.6{\scriptsize $\pm$3.5} & 61.1{\scriptsize $\pm$2.0}
 & 64.0{\scriptsize $\pm$0.9} & 67.9{\scriptsize $\pm$1.2} 
 & 37.7{\scriptsize $\pm$5.4} & 60.4{\scriptsize $\pm$7.8}
 & 73.1{\scriptsize $\pm$6.0} & 95.7{\scriptsize $\pm$1.0} 
 & \textbf{76.4}{\scriptsize $\pm$2.4} & 82.5{\scriptsize $\pm$1.8}
 & 83.9{\scriptsize $\pm$1.4} & 89.0{\scriptsize $\pm$0.4} \\
LLM-Gen 
 & 51.4{\scriptsize $\pm$2.7} & 60.3{\scriptsize $\pm$1.5}
 & 62.7{\scriptsize $\pm$0.9} & 68.5{\scriptsize $\pm$1.1} 
 & 36.6{\scriptsize $\pm$6.6} & 57.3{\scriptsize $\pm$2.7}
 & 71.2{\scriptsize $\pm$4.2} & 95.8{\scriptsize $\pm$0.4} 
 & 69.1{\scriptsize $\pm$6.1} & 78.2{\scriptsize $\pm$2.9}
 & 79.0{\scriptsize $\pm$2.8} & 89.7{\scriptsize $\pm$0.5} \\
LLM-Rew 
 & 49.5{\scriptsize $\pm$3.9} & 60.2{\scriptsize $\pm$1.2}
 & 62.8{\scriptsize $\pm$0.5} & 69.0{\scriptsize $\pm$1.2} 
 & 36.1{\scriptsize $\pm$4.4} & 58.2{\scriptsize $\pm$4.5}
 & 70.8{\scriptsize $\pm$4.0} & 95.7{\scriptsize $\pm$0.4} 
 & 62.4{\scriptsize $\pm$7.0} & 79.3{\scriptsize $\pm$2.5}
 & 82.3{\scriptsize $\pm$1.9} & 89.7{\scriptsize $\pm$0.2} \\
LLM-Mix 
 & 46.9{\scriptsize $\pm$2.5} & 55.4{\scriptsize $\pm$2.2}
 & 60.7{\scriptsize $\pm$1.8} & 68.9{\scriptsize $\pm$0.3} 
 & 31.4{\scriptsize $\pm$6.1} & 57.5{\scriptsize $\pm$7.4}
 & 71.0{\scriptsize $\pm$6.1} & 95.7{\scriptsize $\pm$0.6} 
 & 57.7{\scriptsize $\pm$5.9} & 79.6{\scriptsize $\pm$2.7}
 & 80.6{\scriptsize $\pm$1.5} & \textbf{89.9}{\scriptsize $\pm$0.4} \\
\midrule
inversedMixup 
 & \textbf{52.2}{\scriptsize $\pm$2.5} & \textbf{61.3}{\scriptsize $\pm$1.1}
 & \textbf{64.3}{\scriptsize $\pm$1.3} & \textbf{69.8}{\scriptsize $\pm$0.3} 
 & \textbf{40.9}{\scriptsize $\pm$5.9} & \textbf{61.2}{\scriptsize $\pm$5.8}
 & \textbf{74.4}{\scriptsize $\pm$4.0} & \textbf{96.4}{\scriptsize $\pm$0.4} 
 & 76.3{\scriptsize $\pm$2.8} & 82.5{\scriptsize $\pm$1.5}
 & \textbf{84.6}{\scriptsize $\pm$1.8} & 89.8{\scriptsize $\pm$0.2} \\
\bottomrule
\end{tabular}
\label{tab:qwen_results}
\end{table*}

\begin{table*}
\centering
\caption{Results of LLM-based augmentation methods with ChatGPT-4o.}
\setlength{\tabcolsep}{4.5pt}
\begin{tabular}{c|cccc|cccc|cccc}
\toprule
 & \multicolumn{4}{c|}{Yahoo} & \multicolumn{4}{c|}{TREC} & \multicolumn{4}{c}{AG News} \\
\midrule
 & K=1 & K=5 & K=10 & All & K=1 & K=5 & K=10 & All & K=1 & K=5 & K=10 & All \\
\midrule
LLM-Rew 
 & 50.3{\scriptsize $\pm$2.2} & 60.8{\scriptsize $\pm$1.5} & 63.0{\scriptsize $\pm$1.3} & 69.7{\scriptsize $\pm$1.0}
 & 37.1{\scriptsize $\pm$5.4} & 59.3{\scriptsize $\pm$2.8} & 70.1{\scriptsize $\pm$4.4} & 96.4{\scriptsize $\pm$0.7}
 & 66.3{\scriptsize $\pm$6.0} & 80.4{\scriptsize $\pm$2.8} & 82.1{\scriptsize $\pm$2.3} & 89.7{\scriptsize $\pm$0.1} \\
LLM-Gen 
 & 53.1{\scriptsize $\pm$2.0} & 61.1{\scriptsize $\pm$1.7} & 62.9{\scriptsize $\pm$2.0} & 69.5{\scriptsize $\pm$1.1}
 & 36.8{\scriptsize $\pm$5.8} & 58.4{\scriptsize $\pm$5.5} & 68.9{\scriptsize $\pm$4.3} & 96.2{\scriptsize $\pm$0.6}
 & 65.0{\scriptsize $\pm$6.5} & 81.0{\scriptsize $\pm$2.9} & 81.4{\scriptsize $\pm$2.3} & 89.9{\scriptsize $\pm$0.4} \\
LLM-Mix 
 & 47.2{\scriptsize $\pm$3.3} & 60.5{\scriptsize $\pm$1.3} & 61.6{\scriptsize $\pm$0.7} & 69.6{\scriptsize $\pm$0.8}
 & 35.5{\scriptsize $\pm$7.5} & 59.3{\scriptsize $\pm$7.3} & 68.8{\scriptsize $\pm$4.6} & 95.4{\scriptsize $\pm$1.8}
 & 63.3{\scriptsize $\pm$9.6} & 81.0{\scriptsize $\pm$2.3} & 82.0{\scriptsize $\pm$2.2} & \textbf{90.4}{\scriptsize $\pm$0.8} \\
\midrule
 inversedMixup 
 & \textbf{54.9}{\scriptsize $\pm$2.9} & \textbf{62.1}{\scriptsize $\pm$0.8} & \textbf{65.6}{\scriptsize $\pm$0.7} & \textbf{70.1}{\scriptsize $\pm$0.8}
 & \textbf{38.6}{\scriptsize $\pm$4.6} & \textbf{61.6}{\scriptsize $\pm$6.2} & \textbf{71.2}{\scriptsize $\pm$5.6} & \textbf{96.6}{\scriptsize $\pm$0.9} 
 & \textbf{69.4}{\scriptsize $\pm$6.7} & \textbf{83.4}{\scriptsize $\pm$1.9} & \textbf{84.0}{\scriptsize $\pm$0.9} & 89.7{\scriptsize $\pm$0.9} \\
\bottomrule
\end{tabular}

\label{tab:gpt4o_results}
\end{table*}

\subsection{Effect of Adaptor Refinement}
Alignment and inversion stages are necessary for performing Mixup and making text Mixup observable, rather than optional add-ons. To assess the contribution of the adaptor refinement stage, we compare the full inversedMixup with a variant that omits this stage and instead reconstructs mixed embeddings using only the adaptor aligned on unlabeled data. The results are reported in Table~\ref{tab:ablation}.

As shown in the Table, removing adaptor refinement leads to consistent performance drops across all datasets and settings, confirming that task-specific alignment is essential for inverting mixed embeddings into high-quality augmented samples. Notably, the degradation is most pronounced in low-resource settings (e.g., over 20 points on AG News at K=1) and shrinks substantially as more labeled data becomes available. This indicates that refinement is especially critical when labeled data is scarce.

\subsection{Generalization across Backbone LLMs}
To evaluate whether inversedMixup generalizes across different LLMs,
we fully replace LLaMA-3-8B-Instruct with Qwen2.5-7B-Instruct~\footnote{\url{https://huggingface.co/Qwen/Qwen2.5-7B-Instruct}} as the
backbone for the entire three-stage pipeline. Results are reported in
Table~\ref{tab:qwen_results}.

As shown in the Table, inversedMixup consistently achieves superior or
competitive performance, confirming that its effectiveness does not
depend on a specific LLM architecture.

\subsection{Evaluating with Stronger Labeling LLM}
When mitigating manifold intrusion, inversedMixup relies on an LLM to
assign hard labels, making its performance dependent on the labeling
model's capability. The same applies to other LLM-based augmentation
methods. To examine this effect, we replace the labeling model in
inversedMixup with the stronger
ChatGPT-4o~\footnote{\url{https://openai.com/index/gpt-4o-system-card}},
and likewise adopt ChatGPT-4o for all LLM-based baselines. Results are
reported in Table~\ref{tab:gpt4o_results}.

As shown in the Table, inversedMixup consistently outperforms other LLM-based methods across nearly all settings, demonstrating its effectiveness and robustness even when paired with a more powerful LLM.

\subsection{Effect of the Number of Augmented Samples}

Due to the inherent flexibility of large language models, LLM-based augmentation methods can generate an essentially unlimited number of diverse samples. Similarly, our inversedMixup framework can produce a large volume of samples by varying the interpolation ratios and leveraging the stochastic nature of LLM generation.
To enable a fair comparison of sample quality between the LLM-based approaches, we fix the number of generated samples and evaluate downstream task performance accordingly. The results are presented in Figure~\ref{fig:accurancy_count}.

From the Figure, we observe that inversedMixup consistently outperforms other LLM-based methods, including LLM-Rew, LLM-Gen, and LLM-Mix, across varying numbers of augmented samples. This indicates that, given the same number of samples, inversedMixup not only produces higher-quality samples but also achieves superior augmentation performance, leading to improved results on downstream tasks.

\begin{figure*}
    \centering
    \includegraphics[width=\linewidth]{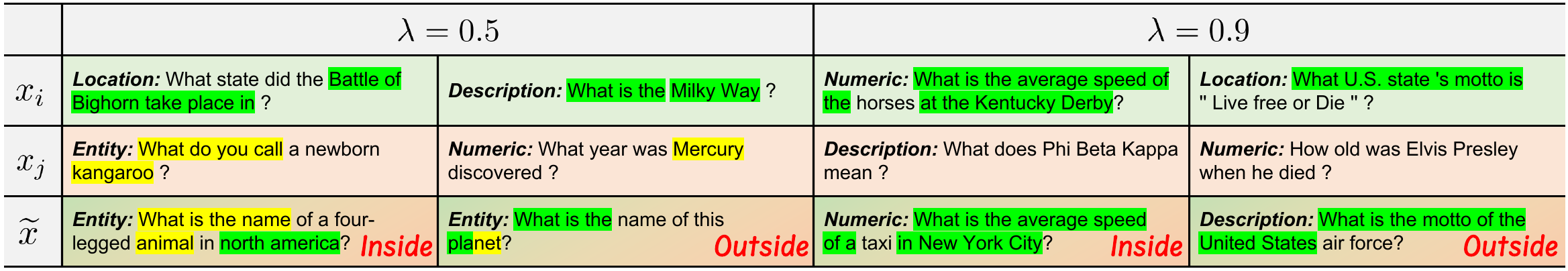}
    \caption{Case study illustrating the manifold intrusion phenomenon under different interpolation ratios ($\lambda = 0.5$ and $\lambda = 0.9$).
    Words highlighted with a green background indicate higher semantic consistency with the label of $x_i$, while those highlighted in yellow are more consistent with the label of $x_j$.
    The `Inside' and `Outside' are consistent with those illustrated in Figure~\ref{fig:intrusion}. 
    }
    \label{fig:case}
\end{figure*}

\subsection{Qualitative Analysis of Manifold Intrusion}
\label{sec:existence}
To better understand the manifold intrusion phenomenon, we conducted a case study by interpolating between two examples using $\lambda = 0.5$ and $\lambda = 0.9$, as shown in Figure~\ref{fig:case}. From the Figure, we observe two key findings:

\textit{Interpolations with ratios near 0.5 are more prone to semantic ambiguity}: When $\lambda = 0.5$, the interpolated sample lies exactly midway between $x_i$ and $x_j$, making it more vulnerable to semantic ambiguity. In contrast, when $\lambda = 0.9$, the interpolation is dominated by one example, which helps maintain semantic coherence. This aligns with intuition: as the interpolation favors one endpoint, the influence of the other decreases.

\textit{Regardless of the interpolation ratio, manifold intrusion can still occur}: Although the interpolated samples $\widetilde{x}$ may appear to semantically blend $x_i$ and $x_j$, their labels often correspond more closely to one of the original categories (`Inside'), rather than representing a soft label. In some cases, the interpolated sample even takes on a completely different label (`Outside'), indicating that the interpolation has drifted into an unrelated manifold. 

Taken together, these findings demonstrate that the intrusion phenomenon is widespread, occurring regardless of whether the synthesized sentence preserves the semantics of the original inputs.

\begin{figure}
    \centering
    \begin{tabular}{c}
        \includegraphics[width=0.47\textwidth]{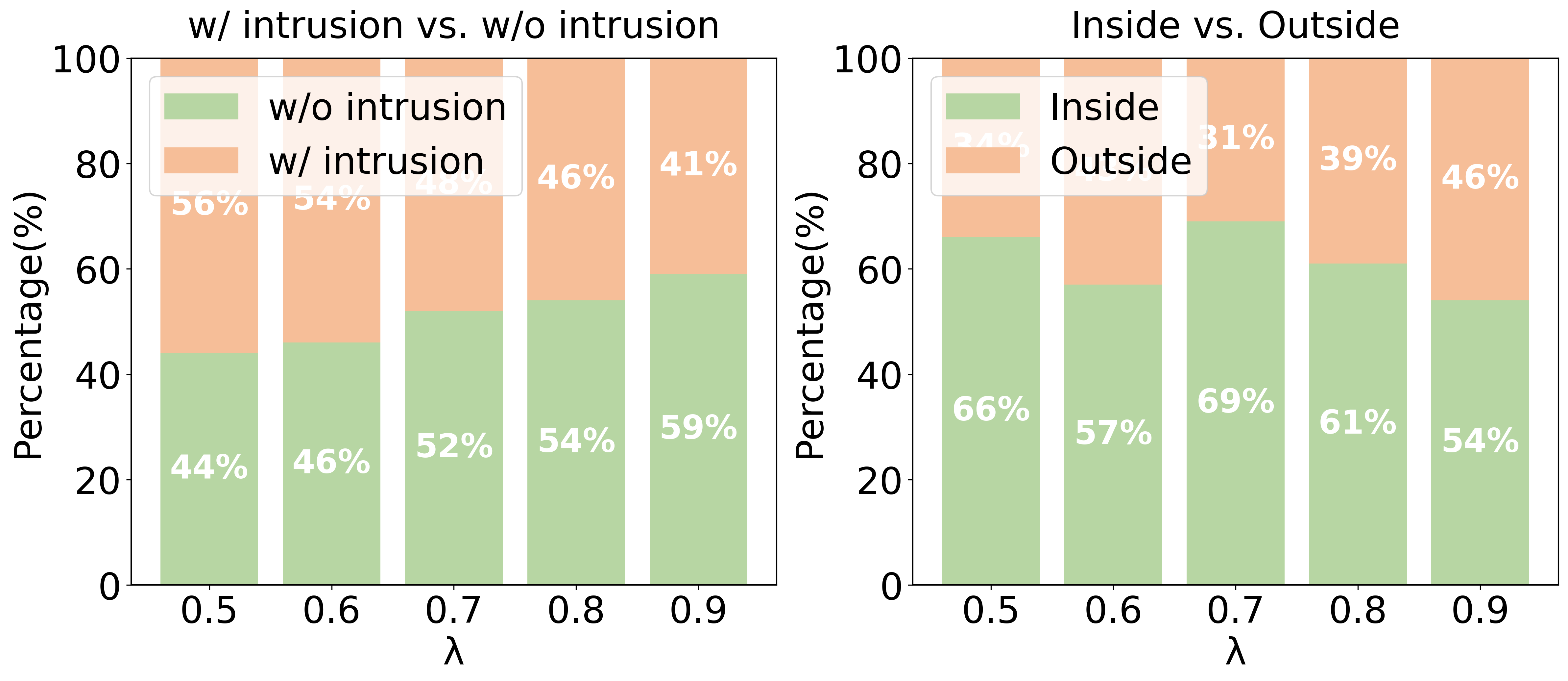}
    \end{tabular}
        \caption{The small figure on the left illustrates the proportion of samples with intrusion (w/ intrusion) versus those without intrusion (w/o intrusion). The right figure shows, among the samples with intrusion, the relative proportions of `Inside' and `Outside' intrusion.}
    \label{fig:intrusion-exp}
\end{figure}

\subsection{Quantitative Analysis of Manifold Intrusion}
After qualitatively confirming the existence of manifold intrusion in NLP tasks, we proceed to quantitatively assess its impact. To this end, we employ an LLM-as-a-Judge~\cite{gu2024survey} approach to determine whether sentences generated by inverting mixed embeddings exhibit intrusion. Specifically, we randomly select 100 augmented sentences from the TREC dataset and use ChatGPT-4o to preliminarily evaluate whether the assigned mixed soft label is appropriate, i.e., whether intrusion occurs. For sentences exhibiting intrusion, we further categorize them based on whether the preferred label falls within the original labels (`Inside') or outside them (`Outside'). The definition of `Inside' and `Outside' refers to Section~\ref{sec:intrusion}. The experimental results are presented in Figure~\ref{fig:intrusion-exp}.

Observing the figure on the left of Figure~\ref{fig:intrusion-exp}, we find that as the mixing ratio increases, the occurrence of intrusion decreases. This aligns with intuition: a larger mixing ratio gives more weight to one of the original sentences, placing the mixed embedding closer to the manifold of the original sentence in the embedding space, which reduces the likelihood of intrusion. Conversely, a smaller mixing ratio increases the chance of intrusion.

Observing the right figure of Figure~\ref{fig:intrusion-exp}, we find that when intrusion occurs, the proportion of `Inside' intrusion is generally higher than that of `Outside'. However, the ratio between `Inside' and `Outside' does not exhibit a linear relationship with the mixing ratio. This may be because the intrinsic distribution of different class manifolds in the embedding space is not always uniform or regular.

\subsection{Alleviation of Manifold Intrusion}
\label{sec:alleviate}
\begin{figure}
    \centering
    \includegraphics[width=0.47\textwidth]{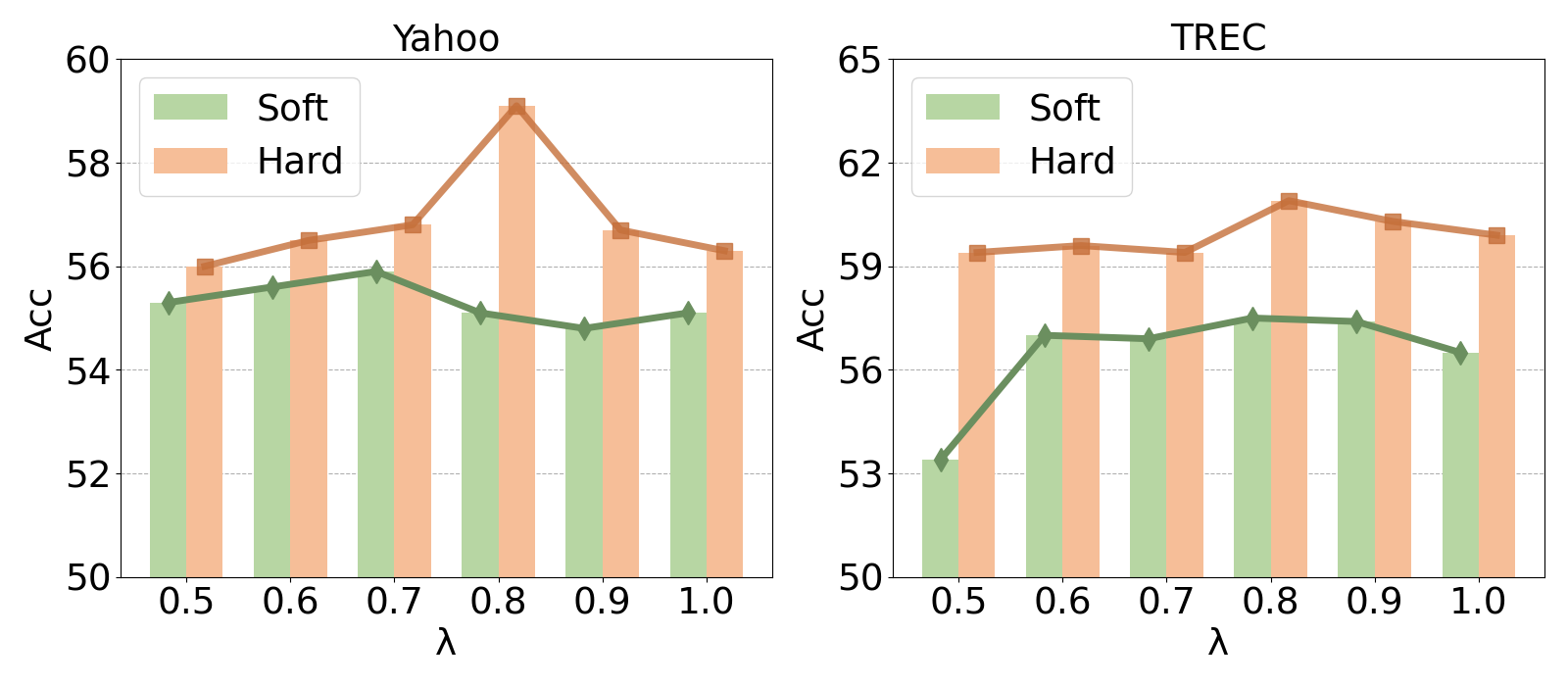}
    \caption{Comparison between soft labels and hard labels for synthesized sentences.}
    \label{fig:soft_hard}
\end{figure}

To alleviate the intrusion, we replace the original interpolated soft labels with hard labels directly assigned by an LLM to the synthetic samples. We compare the performance of hard labels and soft labels under different interpolation ratios $\lambda$, with results shown in Figure~\ref{fig:soft_hard}. Noted, to better observe the experimental outcomes, we fix the interpolation ratio $\lambda$ instead of sampling it from a Beta distribution, which may lead to a slight degradation in overall performance.

From the Figure, we observe that across all $\lambda$ values, hard labels consistently outperform soft labels. This demonstrates that labeling via LLM can alleviate the manifold intrusion phenomenon, helping to stabilize the learning process and better preserve semantic consistency during interpolation.

\section{Conclusion}
In this paper, we present inversedMixup, a unified text augmentation framework that combines the controllability of Mixup with the human-interpretable generation of LLMs. By aligning task-specific embeddings with a frozen LLM through a three-stage training process, our method enables the reconstruction of human-interpretable sentences from interpolated embeddings. This approach not only facilitates the generation of high-quality and semantically coherent synthetic data but also provides a novel means to investigate and mitigate the manifold intrusion phenomenon in text Mixup, offering deeper insights into the relationship between embedding-level manipulations and their effects on downstream task performance.
Extensive experiments demonstrate that inversedMixup consistently improves performance and generalizes effectively across both few-shot and fully supervised scenarios.

Beyond the text domain, the principle of making mixed embeddings
observable could extend to multimodal settings, though challenges
such as cross-modal alignment after interpolation remain to be
addressed. Meanwhile, our hard-label strategy offers a simple
mitigation of manifold intrusion, and a more principled trade-off
between soft and hard labels, especially under outside intrusion,
deserves further investigation.

\begin{acks}
This work was supported by the Postdoctoral Fellowship Program of CPSF under Grant Number GZB20250951, in part by the National Natural Science Foundation of China (No. U2433212), in part by the Fundamental Research Funds for the Central Universities, and in part by the State Key Laboratory of Complex \& Critical Software Environment.
\end{acks}

\bibliographystyle{ACM-Reference-Format}
\bibliography{sample-base}

\appendix

\section{Appendix}

\subsection{Dataset Statistics}
\label{app:data}

Detailed statistics of datasets are shown in Table~\ref{tab:data_stats}. We summarize the number of samples in the training, development, and test sets, as well as the number of classes.
\begin{table}[h]
\centering
\begin{tabular}{ccccc}
\toprule
\textbf{Dataset} & \textbf{\#Train} & \textbf{\#Dev} & \textbf{\#Test} & \textbf{\#Classes} \\
\midrule
Yahoo & 5,000 & 5,000 & 5,000 & 10 \\
TREC & 4,906 & 546 & 500 & 6 \\
AG News & 5,000 & 4,999 & 2,600 & 4 \\
\bottomrule
\end{tabular}
\caption{Dataset statistics.}
\label{tab:data_stats}
\end{table}

For our few-shot experiments, we randomly sampled $K$-shot examples ($K=1, 5, 10$) from the training set.

\subsection{Baselines}
\label{app:baseline}

For the baseline methods, we re-implemented the models to ensure a fair comparison. Details of the baseline implementations are provided below.

\noindent\underline{\textbf{EDA}}~\cite{wei2019eda} is a simple yet effective technique that generates diverse training examples using four operations: synonym replacement, random insertion, random swap, and random deletion.

\noindent\underline{\textbf{BT}}~\cite{sennrich2015improving} uses pre-trained translation models, generating paraphrased examples for each original sentence.

\noindent\underline{\textbf{Textsmooth}}~\cite{wu2022text} replaces one-hot token representations with smoothed distributions from a masked language model.

\noindent\underline{\textbf{AWD}}~\cite{chen2023adversarial} generates hard positive examples by weakening key class-indicative words through adversarial embedding mixing.

\noindent\underline{\textbf{Mixup}}~\cite{zhang2017mixup} linearly interpolates sentence embeddings and corresponding labels.

\noindent \underline{\textbf{LLM-Rew}} encourages the model to produce semantically equivalent yet lexically diverse reformulations.
\begin{promptbox}
Please provide a clear and concise rewrite of this \{label\} sentence: ``\{text\}"

Rewritten version:
\end{promptbox}

\noindent \underline{\textbf{LLM-Gen}} enhances dataset diversity by generating lexically and syntactically varied sentences while preserving the original labels.
\begin{promptbox}
Write a \{label\} \{topic\} sentence. Output only the sentence, nothing else.

The sentence is:
\end{promptbox}

\noindent \underline{\textbf{LLM-Mix}}  encourages the model to generate a coherent and concise fusion of both inputs while preserving their combined meaning.
\begin{promptbox}
Mix the two sentences into one sentence:

sentence1: \{text1\}

sentence2: \{text2\}

The mixed sentence is:
\end{promptbox}

\subsection{Utilizing LLM for Labeling}
To facilitate automatic labeling using large language models (LLMs), we carefully design a set of prompt templates that guide the model to generate accurate and label-consistent annotations.

\begin{promptbox}
Classify the following sentence into one of the labels below.
Sentence: ``\{text\}"

Labels:
\{label\_list\}

Respond with only the label name exactly as given above.
Answer:
\end{promptbox}

To promote accurate labeling, we accompany each label with a brief description to clarify its intended semantics.
\begin{promptbox}
        ``Description": ``descriptions of objects, people, or events",
        
        ``Numeric": ``numerical data such as dates, amounts, or quantities",
        
        ``Entity": ``specific named entities like objects, products, or organizations",
        
        ``Location": ``geographical places, cities, countries, or regions",
        
        ``Human": ``people, occupations, or human-related concepts",
        
        ``Abbreviation": ``shortened forms of words or acronyms",
        
        ``World": ``topics related to international events, diplomacy, or global issues",
        
        ``Sports": ``sports news, teams, players, or competitions",
        
        ``Business": ``economic topics such as finance, companies, and markets",
        
        ``Sci/Tech": ``science, technology, innovation, or research topics",
        
        ``Entertainment": ``topics related to movies, music, celebrities, or pop culture",
        
        ``Relationship": ``personal relationships, dating, or social interactions",
        
        ``Technology": ``technological tools, trends, and advancements",
        
        ``Health": ``healthcare, medicine, wellness, or diseases",
        
        ``Science": ``scientific research, discoveries, and theories",
        
        ``Finance": ``financial topics such as banking, investing, or economics",
        
        ``Culture": ``arts, traditions, beliefs, or societal norms",
        
        ``Politics": ``government, political debates, or policy-making",
        
        ``Education": ``schools, learning systems, or educational content"
\end{promptbox}

\subsection{Quantitative Analysis of Manifold Intrusion through LLM}

To precisely examine the intrusion phenomenon, we design two prompt sets for the large model: one to determine whether intrusion occurs and another to specify its type, i.e., `Inside' or `Outside' the original label space.

The first set of instructions is used to assess the validity of the interpolated soft labels with respect to the ground truth.

\begin{promptbox}
You are a text classification model.

Determine if the sentence "\{text\}" can be labeled as 

"\{label\_text\}".

If it matches, output "Yes"; otherwise, output "No".
\end{promptbox}

The second set of prompts instructs the model to assign a categorical label, thereby determining whether the observed intrusion should be classified as `Inside' or `Outside'.

\begin{promptbox}
You are a text classification model.

Choose one label for the sentence "\{text\}" in "\{label\_map\}".
\end{promptbox}

\end{document}